\pdfoutput=1

\documentclass[11pt]{article}

\usepackage[preprint]{acl}
\usepackage{amsmath}

\usepackage{times}
\usepackage{latexsym}
\usepackage{arydshln}
\usepackage[T1]{fontenc}

\usepackage[utf8]{inputenc}

\usepackage{microtype}

\usepackage{inconsolata}

\usepackage{graphicx}

\title{Extrapolation Merging: Keep Improving With
Extrapolation and Merging}

\author{Yiguan Lin\thanks{~~These authors are both First Author.}~~~Bin Xu$^{*}$~~~Yinghao Li~~~Yang Gao\thanks{~~Corresponding author} \\
School of Computer Science and Technology, Beijing Institute of Technology, Beijing, China \\
\texttt{\{yglin,binxu,yhli,gyang\}@bit.edu.cn}
}

\begin{document}
\maketitle
\begin{abstract}


Large Language Models (LLMs) require instruction fine-tuning to perform different downstream tasks. However, the instruction fine-tuning phase still demands significant computational resources and labeled data, lacking a paradigm that can improve model performance without additional computational power and data. Model merging aims to enhance performance by combining the parameters of different models, but the lack of a clear optimization direction during the merging process does not always guarantee improved performance. In this paper, we attempt to provide a clear optimization direction for model merging. We first validate the effectiveness of the model extrapolation method during the instruction fine-tuning phase. Then, we propose Extrapolation Merging, a paradigm that can continue improving model performance without requiring extra computational resources or data. Using the extrapolation method, we provide a clear direction for model merging, achieving local optimization search, and consequently enhancing the merged model's performance. We conduct experiments on seven different tasks, and the results show that our method can consistently improve the model's performance after fine-tuning.
\end{abstract}

\begin{figure}[!bt]
    \centering
    \includegraphics[width=0.85\linewidth]{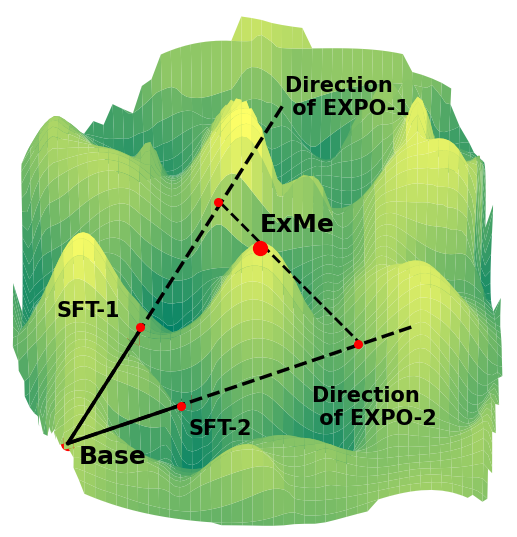}
    \caption{Overview of ExMe: ExMe process begins with a base model, from which SFT is conducted and the top two checkpoints with the best overall performance are selected. These SFT models are then extrapolated independently from the base model, resulting in two extrapolated models, EXPO-1 and EXPO-2. Finally, the two extrapolated models are directly weighted and merged, yielding the final merged model. In this figure, the color depth of the surface represents the model's performance, with lighter colors indicating stronger performance.}
    \label{fig:Final_mountain}
\end{figure}

\section{Introduction}


Large language models (LLMs) typically require an alignment phase to effectively understand and follow human instructions~\cite{DBLP:journals/corr/abs-2303-18223}. This phase often involves Supervised Fine-Tuning (SFT) and Reinforcement Learning from Human Feedback (RLHF)~\cite{DBLP:conf/iclr/WeiBZGYLDDL22,DBLP:conf/nips/Ouyang0JAWMZASR22}, both of which demand high-quality data for downstream tasks and significant computational resources~\cite{DBLP:journals/corr/abs-2001-08361,DBLP:journals/corr/abs-2203-15556}. As a result, the current alignment process for large models lacks a paradigm that does not require additional data and computational power.

Recently, studies on \textit{Model Merging} have demonstrated that it is possible to enhance model performance by simply merging weight, without the need for additional data or computational power~\cite{DBLP:conf/iclr/Jin0P023,DBLP:conf/iclr/IlharcoRWSHF23}. 
This offers a potential paradigm for model alignment. However, most current model merging strategies primarily emphasize hyperparameter search and the identification of redundant parameters~\cite{DBLP:conf/nips/YadavTCRB23,DBLP:journals/corr/abs-2311-03099}, with the performance of the merged model typically falling between the models being merged~\cite{lin2024mitigatingalignmenttaxrlhf}. These challenges create ambiguity in the merging process, leading to unpredictable outcomes in the merged model's performance. Therefore, establishing a well-defined strategy for merging model weights is of critical importance.

\textit{Model Extrapolation} (\textbf{EXPO}) ~\cite{DBLP:journals/corr/abs-2404-16792} is a method derived from model merging that similarly enhances model performance by adjusting parameter matrix without requiring additional data or training, offering a promising optimization direction for model merging. By employing an inverse strategy to model merging, model extrapolation views DPO/RLHF~\cite{DBLP:conf/nips/RafailovSMMEF23} models as the outcome of merging an SFT model with an unknown but more powerful model (EXPO Model), thereby enabling the inference of EXPO model's parameters. Therefore, the higher-performing models obtained through extrapolation can serve as the foundational starting points for model merging. However, determining how to select the appropriate model checkpoints using the extrapolation method remains a challenge that needs to be addressed.

However, model extrapolation only optimizes on a single direction of model optimization, without considering the potential multiple optimization paths of the model during training. If only a model checkpoint from a single optimization path is selected for model merging, the model is prone to getting stuck in a local optimum region. Furthermore, since the DPO/RLHF training phase uses a model that has already been through SFT, the model parameters after this phase are already in a relatively optimal position, leading to a narrower optimization direction for DPO/RLHF on this model. Contrastly, if extrapolation is employed based on the base model during the SFT phase and multiple optimization paths are considered, it would provide a broader optimization space for the model and enable model merging based on these optimization paths, thereby stabilizing the achievement of a model with better performance

In this paper, we propose a paradigm for model alignment that can be achieved without additional data or computational power, called \textbf{Ex}trapolation \textbf{Me}rging (\textbf{ExMe}), which integrates model extrapolation and merging methods. This approach enhances model performance comprehensively by adjusting model parameters alone. As illustrated in Figure~\ref{fig:Final_mountain}, we apply extrapolation during the SFT stage to generate models optimized in different directions, and then merge these models. Specifically, we begin by fine-tuning the base model to obtain multiple SFT models, which are considered the result of merging the base model with an unknown but more powerful model. By applying the extrapolation method during the SFT stage, we obtain models with superior performance, thereby validating the effectiveness of the extrapolation approach on the SFT stage. Furthermore, after determining multiple optimization directions using the extrapolation method, we select the two best-performing models as the starting points for model merging. This approach yields a merged model with higher comprehensive performance across different optimization directions. Our contributions are as follows:
\begin{itemize}
    \item We validate the effectiveness of model extrapolation on the SFT stage.
    \item We provide a clear and effective optimization direction for model merging.
    \item We propose a simple new paradigm for model alignment that does not require additional computational power or data, ExMe, that enhances model performance by adjusting model weight without relying on additional fine-tuning data or training.
\end{itemize}

\begin{figure*}[!htbp]
    \centering
    \includegraphics[width=0.85\textwidth]{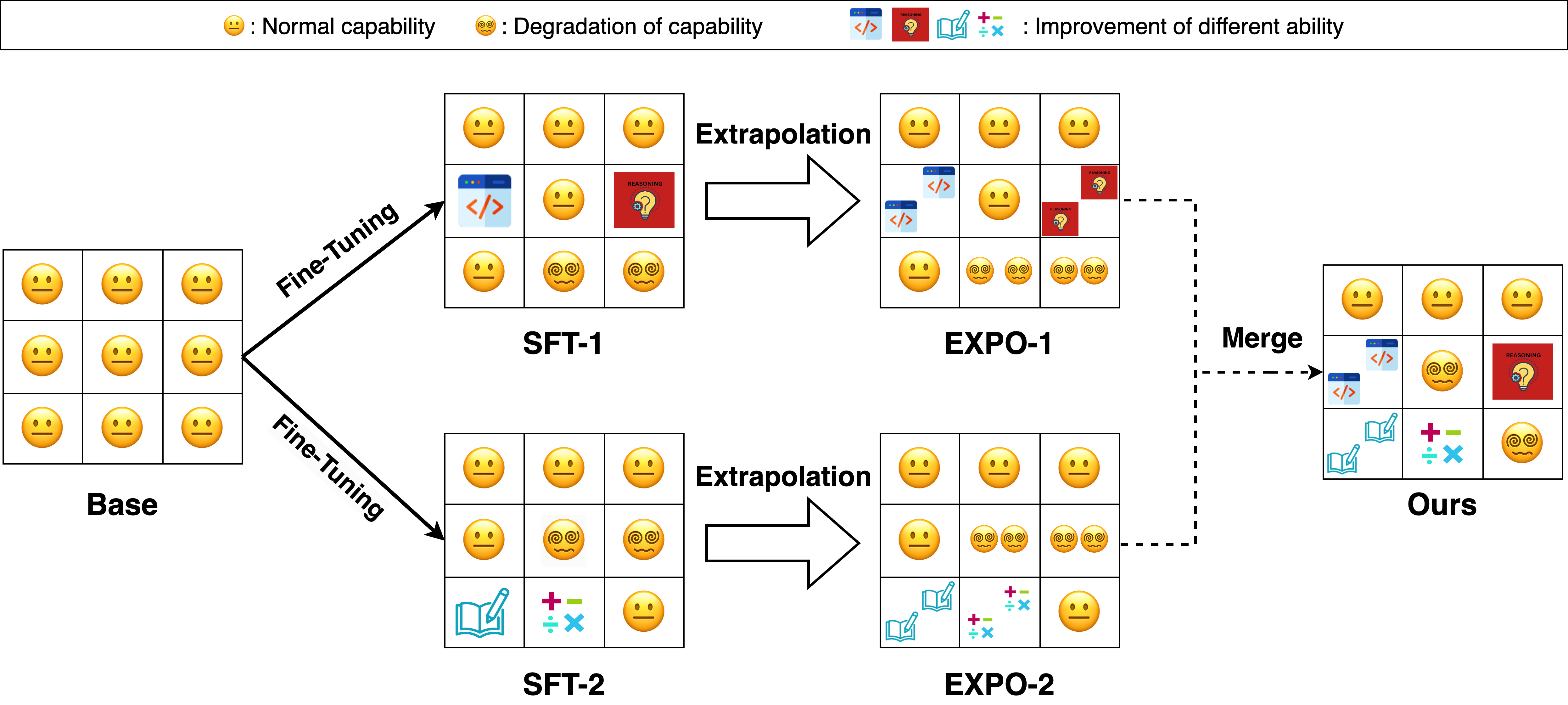}
    \caption{Three stages of ExMe: In the fine-tuning stage, after the base model is fine-tuned, a series of checkpoints are generated, from which the two SFT models with the best overall performance are selected for the next stage of extrapolation. In the extrapolation stage, each of these SFT models emphasizes different capabilities, which are further amplified in the best two directions. In the merging stage, the merging process balances the strengths and mitigates the weaknesses of the two extrapolated models, resulting in improved overall performance.}
    \label{fig:pipeline}
\end{figure*}

\section{Related Work}
\subsection{Supervised Fine-Tuning of LLMs}
Supervised Fine-Tuning (SFT) fine-tune the pre-trained LLMs on a labeled dataset, utilizing labeled datasets with task-specific input-output pairs \cite{DBLP:journals/corr/abs-2210-11416}. This process enhances the model's ability to generate accurate and contextually relevant responses tailored to the specific task, such as question-answering, summarization, and translation.
However, LoRA \cite{DBLP:conf/iclr/HuSWALWWC22} discovered that when LLMs are adapted for specific tasks, the parameter matrices are often over-parameterized, with typically high dimensions but a low intrinsic rank. To address this issue, LoRA proposes integrating low-rank decomposition matrices into the parameter matrices of pre-trained LLMs to approximate parameter updates at each layer, thereby reducing the number of parameters needed to fine-tune downstream tasks. Compared to full-parameter fine-tuning, LoRA significantly reduces the computational cost while maintaining model performance, and has been widely adopted for parameter-efficient fine-tuning of open-source LLMs.

\subsection{Model Merging and Model Extrapolation}
Model merging has emerged as a research trend in recent years, aiming to leverage existing models with different capabilities to efficiently create a unified model with diverse capabilities by adjusting weight parameters. 
Model averaging~\cite{DBLP:conf/icml/WortsmanIGRLMNF22} is a common method for model merging, constructing the merged model by averaging the weight parameters of different models. Weighted merging assigns different weights to the models involved, with Fisher-merging employing the Fisher information matrix to determine the optimal merging hyperparameters~\cite{DBLP:conf/nips/MatenaR22}. Bayesian search~\cite{DBLP:journals/corr/abs-2403-19390} begins with a series of pre-trained checkpoints and uses Bayesian optimization to find the optimal merging hyperparameters. 
MoLE~\cite{DBLP:conf/iclr/WuHW24} approaches merging within a single model, treating each trained LoRA as an independent expert, controlling hierarchical weights by integrating a learnable gating function within each layer to learn the optimal weight combinations for different objectives.
DARE~\cite{DBLP:journals/corr/abs-2311-03099} points out that the delta parameters of various SFT models (the disparity between SFT and pre-trained parameters) are highly redundant, and thus employs DARE to eliminate redundant incremental parameters in each model before merging them.
However, these model merging methods often overlook the optimization direction of parameters.

Model extrapolation, derived from weighted merging, adopts a reverse approach by treating the DPO/RLHF model as an interpolated result between an SFT model and a stronger model, thereby facilitating the extrapolation of a more powerful model from the SFT model and the existing DPO/RLHF model.

Unlike traditional model merging methods, model extrapolation considers the optimization direction of parameters, using human preference data to guide parameter adjustments and linearly amplifying the effects of reinforcement learning on these adjustments. However, model extrapolation has so far demonstrated effectiveness primarily between SFT models and DPO/RLHF models.

\section{Method}

In this section, we first introduce the mechanisms of model merging and extrapolation methods. We then conclude by presenting our ExMe method, which is derived from these two approaches.

\subsection{Model Merging} 
Model merging aims to utilize multiple distinct model checkpoints by applying mathematical operations, such as weighted merging, to merge them into a unified model that consolidates the strengths of each. This technique is capable of efficiently enhancing model performance.
Formally, let ${\Theta_1, \Theta_2, \ldots, \Theta_t}$ represent $t$ checkpoints with the same architecture. The linear combination of these model checkpoints in the parameter space can be expressed in the following form, which is a general weighted merging formula:
\begin{equation} 
\label{eq:1}
{\Theta}_{merged} = \sum_{i=1}^{t} \lambda_i \Theta_i \quad \text{s.t.} \quad \sum_{i=1}^{t} \lambda_i = 1
\end{equation}
where $\lambda_i \in [0, 1]$ represents the weight of the $i$-th model checkpoint involved in the merging process, and $\Theta_{merged}$ represents the merged model resulting from the merging process.

Merging more models does not necessarily improve the performance of the merged model. Previous studies have shown that when merging checkpoints from models with the same architecture, the performance of the merged model often declines as the number of models involved increases. Typically, merging two models yields the best results~\cite{DBLP:journals/corr/abs-2403-19390}. Therefore, this paper employs a strategy of merging only two checkpoints, and equation~\ref{eq:1} can be rewritten as:
\begin{equation}
\label{eq:2}
{\Theta}_{merged} = \lambda _{} \Theta _{1} +  \left ( 1- \lambda _{}  \right ) \Theta _{2}
\end{equation}
where $\lambda \in [0, 1]$ represents the weight of $\Theta_1$, and $1-\lambda$ represents the weight of $\Theta_2$ in the merging process.

\subsection{Model Extrapolation} 
Model extrapolation is a variant of model merging techniques that exploits parameter trends between a weaker and a stronger model to extrapolate a more powerful model.
Formally, let $\Theta_{weak}$ represents the weaker model and $\Theta_{strong}$ represents the stronger model.
The stronger model $\Theta_{strong}$ can be seen as the result of merging $\Theta_{weak}$ with an unknown but more powerful model, denoted as $\Theta_{EXPO}$, according to Equation 2 that can be rewritten as:
\begin{equation} 
\label{eq:3}
{\Theta}_{strong}  = \lambda _{} \Theta _{EXPO} +  \left ( 1- \lambda _{}  \right ) \Theta _{weak}
\end{equation}
Then by substituting equation~\ref{eq:3}, we can obtain:
\begin{align} 
\Theta _{EXPO} &=  \frac{  \Theta _{strong} -  \left ( 1- \lambda _{}  \right ) \Theta _{weak}   }{\lambda _{}} \label{eq:4}
\\ &= \frac{1}{\lambda _{}}\Theta _{strong} - \frac{1-\lambda _{}}{\lambda _{}} \Theta _{weak}  \label{eq:5}
\end{align} 
 Let the extrapolation hyperparameter $\alpha = \frac{1}{\lambda} - 1$, then equation~\ref{eq:5} can be further simplified to the following form:
\begin{equation} 
\label{eq:6}
\Theta _{EXPO} = \Theta _{strong} + \alpha \left ( \Theta _{strong} - \Theta _{weak}\right ) 
\end{equation}

\begin{table*}[h]
\centering
\resizebox{0.87\textwidth}{!}{%
\begin{tabular}{cccccccc}
\hline
\multicolumn{1}{c|}{Model} &
\multicolumn{1}{c|}{\begin{tabular}[c]{@{}c@{}}Optimal \\ hyperparameters\end{tabular}} & 
\multicolumn{1}{c|}{\begin{tabular}[c]{@{}c@{}}CMMLU\\ 5-shot\end{tabular}} &
\multicolumn{1}{c|}{\begin{tabular}[c]{@{}c@{}}C-Eval\\ 5-shot\end{tabular}} &
\multicolumn{1}{c|}{\begin{tabular}[c]{@{}c@{}}HumanEval\\pass@1\end{tabular}} &
\multicolumn{1}{c|}{GSM8K} &
\multicolumn{1}{c|}{GaoKaoBench} &
\multicolumn{1}{c}{Average} \\ \hline
\multicolumn{8}{c}{\textbf{Qwen2-7B}} \\ \hline
\multicolumn{1}{l}{Base} & -   &  \textbf{83.10} &  82.71 &  56.10 &  22.29 &  51.57 &  59.15 \\   \hdashline
\multicolumn{1}{l}{SFT-1} &  - &  80.68          &  82.78 &  67.68 &  41.62 &  65.21 &  67.60 \\
\multicolumn{1}{l}{SFT-2} &  - &  80.77*         &  83.38*&  67.68*&  42.76*&  65.92*&  68.10*\\   \hdashline
\multicolumn{1}{l}{Weighted Merging} & $\lambda$ = 0.2  &  80.76 &  \textbf{83.46} & 67.68 &  43.29 &  65.97 &  68.23 \\   
\multicolumn{1}{l}{Ties} &  - &  80.75     & 83.54 & 67.68 & 42.68 & 66.04 & 68.14 \\
\multicolumn{1}{l}{DARE} &  - &   80.81       &83.33  &67.68  &41.02  & 66.03	& 67.77 \\   \hdashline
\multicolumn{1}{l}{EXPO-1} &  $\alpha$ = 0.5  &  79.93 &  82.44 &  67.68 &  58.26 &  66.29 &  70.92  \\
\multicolumn{1}{l}{EXPO-2} &  $\alpha$ = 0.1 &  80.54 &  83.27 &  67.68 &  46.40 &  \textbf{66.96} &  68.97  \\
\multicolumn{1}{l}{ExMe} &    $\beta$ = 0.1  &  79.90 &  82.56 &  \textbf{67.68} &  \textbf{60.80} &  65.71 &  \textbf{71.33} \\ \hline
\multicolumn{8}{c}{\textbf{Qwen1.5-14B}} \\ \hline
\multicolumn{1}{l}{Base} &    -  &  \textbf{76.53} &  76.68 &  63.41 &  61.18 &  44.22 &  64.40 \\    \hdashline
\multicolumn{1}{l}{SFT-1} &    -  &  76.03* &  77.05* &  64.02* &  63.38* &  67.60 &  69.61* \\
\multicolumn{1}{l}{SFT-2} &    -  &  75.83 &  76.89 &  60.36 &  63.31 &  70.29* &  69.33 \\    \hdashline
\multicolumn{1}{l}{Weighted Merging} &    $\lambda$ = 0.4  &  75.93 &  77.07 &  63.46 &  63.46 &  68.41 &  69.67 \\    
\multicolumn{1}{l}{Ties} &  - &    76.01      & 77.15 & 62.20 & 63.76 & 63.76 & 68.58 \\
\multicolumn{1}{l}{DARE} &  - &     76.04     & 77.47 &  61.59 & 63.91 & 67.49 & 69.30 \\   \hdashline
\multicolumn{1}{l}{EXPO-1} &    $\alpha$ = 0.4  &  75.44 &  77.27 &  60.36 &  64.14 &  \textbf{80.42} &  71.52 \\
\multicolumn{1}{l}{EXPO-2} &    $\alpha$ = 0.3  &  75.57 &  77.14 &  62.19 &  63.76 &  80.16 &  71.76 \\
\multicolumn{1}{l}{ExMe} &    $\beta$ = 0.1  &  75.46 &  \textbf{77.39} &  \textbf{64.63} &  \textbf{64.82} &  80.08 &  \textbf{72.48} \\ \hline
\multicolumn{7}{c}{} \\ \hline

\end{tabular}%
}
\caption{This table shows the scores of Qwen2-7B, Qwen1.5-14B using different methods. * indicates the highest individual score among the two SFT models.}

\label{tab:ComprehensiveComparison1}
\end{table*}

\begin{table*}[h]
\centering
\resizebox{0.87\textwidth}{!}{%
\begin{tabular}{cccccccc}
\hline
\multicolumn{1}{c|}{Model} & 
\multicolumn{1}{c|}{\begin{tabular}[c]{@{}c@{}}Optimal \\ hyperparameters\end{tabular}} &  
\multicolumn{1}{c|}{MMLU} & \multicolumn{1}{c|}{MATH} & \multicolumn{1}{c|}{\begin{tabular}[c]{@{}c@{}}HumanEval\\pass@1\end{tabular}} & \multicolumn{1}{c|}{GSM8K} & \multicolumn{1}{c|}{GaoKaoBench} & \multicolumn{1}{c}{Average} \\ \hline
\multicolumn{8}{c}{\textbf{Meta-Llama-3-8B}} \\ \hline
\multicolumn{1}{l}{Base} &    -  & \textbf{64.9} & 13.42 & 33.50 & 56.03 & 20.24 & 37.62 \\
\hdashline
\multicolumn{1}{l}{SFT-1} &    -  & 64.30 & 16.02* & 38.41 & 60.58 & 24.57 & 40.78 \\
\multicolumn{1}{l}{SFT-2} &    -  & 64.50* & 15.48 & 41.46* & 60.73* & 25.82* & 41.60* \\   \hdashline
\multicolumn{1}{l}{Weighted Merging} &    $\lambda = 0.1$  & 64.40 & 15.62 & 44.51 & 60.73 & 24.82 & 42.02 \\
\multicolumn{1}{l}{Ties} &  - &    64.50      & 15.46 & 43.29 & 60.73 & 25.69 &  41.93\\
\multicolumn{1}{l}{DARE} &  - &    64.40      & 15.58 &43.90  & 60.12 & 26.08 &  42.02\\   \hdashline
\multicolumn{1}{l}{EXPO-1} &    $\alpha = 0.3$  &63.90 & \textbf{16.24} & 45.73 & 60.58 & 24.53 & 42.20 \\
\multicolumn{1}{l}{EXPO-2} &    $\alpha$ = 0.4  &63.90 & 15.40 & 45.12 & \textbf{61.17} & 24.69 & 42.06 \\
\multicolumn{1}{l}{ExMe} & $\beta$ = 0.2  & 63.90 & 16.16 & \textbf{46.95} & 60.73 & \textbf{26.39} & \textbf{42.83}\\ \hline
\multicolumn{8}{c}{\textbf{Mistral-Nemo-Base-2407}} \\ \hline
\multicolumn{1}{l}{Base} &     -  &63.10 & 16.66 & 50.61 & 58.76 & 21.80 & 42.19 \\
\hdashline
\multicolumn{1}{l}{SFT-1} &     -  &64.10 & 16.46* & 53.66 & \textbf{66.19}* & 42.51* & 48.58* \\
\multicolumn{1}{l}{SFT-2} &     -  &64.30* & 15.92 & 54.27* & 66.03 & 42.14 & 45.53 \\   \hdashline
\multicolumn{1}{l}{Weighted Merging} &  $\lambda = 0.2$  &64.50  & 16.34 &53.05  & 66.19 & 42.58 & 48.53 \\
\multicolumn{1}{l}{Ties} &  - &   64.30       & 15.94 & 48.17 & 66.26 & 43.31 &  47.60\\
\multicolumn{1}{l}{DARE} &  - &     64.20     & 15.64 & 49.39 & 66.94 & 43.94 & 48.02 \\   \hdashline
\multicolumn{1}{l}{EXPO-1} &   $\alpha$ = 0.5  &64.50 & 16.26 & 56.10 & 65.28 & 41.82 & 48.79 \\
\multicolumn{1}{l}{EXPO-2} &  $\alpha$ = 0.1  &64.40 & 15.70 & 54.27 & 65.50 & 42.20 & 48.41 \\
\multicolumn{1}{l}{ExMe} &   $\beta$ = 0.1 &\textbf{64.50} & \textbf{16.78} & \textbf{56.10} & 65.43 & \textbf{42.63} & \textbf{49.09} \\ \hline
\end{tabular}%
}
\caption{This table shows the scores of Meta-Llama-3-8B, and Mistral-Nemo-Base-2407 using different methods. * indicates the highest individual score among the two SFT models.}

\label{tab:ComprehensiveComparison2}
\end{table*}

\subsection{ExMe}

ExMe is a method that guides parameter optimization through extrapolation and employs standard model merging techniques to refine the extrapolated model.

ExMe begins by fine-tuning the base model to obtain a series of SFT models, denoted as $\Theta_{ckpt_1}$ to $\Theta_{ckpt_n}$. Next, a comprehensive evaluation of the saved SFT models is conducted, and the top two models with the best overall capabilities are selected, denoted as $\Theta_{SFT-1}$ and $\Theta_{SFT-2}$. 

Next, by applying the model extrapolation on SFT stage, we treat the SFT model and base model as the aforementioned strong and weak model, respectively. According to Equation~\ref{eq:6}, we can obtain several extrapolated models by manually setting different extrapolation hyperparameters. Then we select two extrapolated models with the best overall performance, denoted as $\Theta_{EXPO-1}$ and $\Theta_{EXPO-2}$, as shown in the following equations:
\begin{subequations}
\begin{align}
\Theta_{EXPO-1} &= \scriptstyle \Theta_{SFT-1} + \alpha_{1} \left ( \Theta_{SFT-1} - \Theta_{base} \right ) \tag{\textit{7a}} \label{eq:7a} \\
\Theta_{EXPO-2} &= \scriptstyle \Theta_{SFT-2} + \alpha_{2} \left ( \Theta_{SFT-2} - \Theta_{base} \right ) \tag{\textit{7b}} \label{eq:7b}
\end{align}
\end{subequations}

Finally, by adjusting various merging hyperparameter $\beta$, we merge $\Theta_{EXPO-1}$ and $\Theta_{EXPO-2}$ according to Equation~\ref{eq:2}, leading to several candidate models, which is formulated as:
\begin{equation} 
\label{eq:8}
{\Theta }_{ExMe} = \beta _{} \Theta _{EXPO-1} +  \left ( 1- \beta  _{}  \right ) \Theta _{EXPO-2}
\end{equation}
where $\beta\in[0, 1]$ represents the weight of $\Theta _{EXPO-1}$, and $1-\beta$ represents the weight of $\Theta _{EXPO-2}$ in the merging process.
After evaluating these candidate models, we select the best performing one as the final ExMe model.

\section{Experiments}
\subsection{Pipeline}

We present the pipeline of the ExMe method in Figure~\ref{fig:pipeline}, which consists of three stages: Supervised Fine-Tuning, Extrapolation, and Merging. 
\paragraph{SFT Stage}We first fine-tune the base model using an instruction-following dataset, then evaluate a series of SFT models and select the top two models based on their overall performance. As shown in Figure~\ref{fig:pipeline}, SFT-1 and SFT-2 represent the selected models.
\paragraph{Extrapolation Stage} Based on equations~\ref{eq:7a} and~\ref{eq:7b}, the selected SFT models, SFT-1 and SFT-2, are extrapolated with the base model using manually set values of $\alpha \in \{0.1, 0.2, \dots, 0.5\}$. The two extrapolated models with the best overall performance, shown in Figure~\ref{fig:pipeline} as Extrapolated-1 and Extrapolated-2, are then selected to be utilized in the model merging stage.
\paragraph{Merging Stage} Finally, based on equation~\ref{eq:8}, we merge the two selected extrapolated models with the manually set merging hyperparameter $\beta \in \{0.1, 0.2, \dots, 0.9\}$. Then we evaluate these merged models to select the one with the best overall performance as the final ExMe model.

\subsection{Models} 
We conduct test experiments across models of different scales, specifically 7B, 8B, 12B, and 14B. The following representative models are selected as base models for our experiments: Qwen2-7B~\cite{qwen2}, Meta-Llama-3-8B~\cite{llama3modelcard}, Mistral-Nemo-Base-2407~\cite{mistral-nemo}, Qwen1.5-14B~\cite{qwen1.5}.

\subsection{Datasets}
We use alpaca-gpt4-data-zh and alpaca-gpt4-data-en~\cite{peng2023instruction} as the SFT dataset. These datasets, generated by GPT-4, contain 52K high-quality instruction data. 
For the Chinese LLMs (Qwen2-7B and Qwen1.5-14B), we use alpaca\_gpt4\_data\_zh for SFT and for the English LLMs (Meta-Llama-3-8B and Mistral-Nemo-Base-2407), we use alpaca\_gpt4\_data\_en.

\subsection{Evaluation}

We evaluate the capabilities of the LLMs using OpenCompass~\cite{2023opencompass}, an integrated evaluation framework that includes various mainstream evaluation datasets. In the three stages of ExMe, we use the same dataset to evaluate LLMs. To conduct a more comprehensive evaluation of the capabilities of these models, we select the following seven benchmark datasets:
\textbf{GSM8K}~\cite{DBLP:journals/corr/abs-2110-14168};
\textbf{MATH}~\cite{DBLP:conf/nips/HendrycksBKABTS21};
\textbf{HumanEval}~\cite{DBLP:journals/corr/abs-2107-03374};
\textbf{MMLU}~\cite{DBLP:conf/iclr/HendrycksBBZMSS21};
\textbf{CMMLU}~\cite{DBLP:journals/corr/abs-2306-09212};
\textbf{C-Eval}~\cite{DBLP:conf/nips/HuangBZZZSLLZLF23};
\textbf{GaoKaoBench}~\cite{Zhang2023EvaluatingTP}



\section{Results and Analysis}

\subsection{Overview}

The main results are shown in Table~\ref{tab:ComprehensiveComparison1} and Table~\ref{tab:ComprehensiveComparison2}, which presents the scores of four base models evaluated across multiple datasets after undergoing SFT, Weighted Merging, Ties-Merging, DARE-Merging, Model Extrapolation, and Extrapolation Merging respectively.

\paragraph{Compared with other merging methods, ExMe has the greatest improvement in the comprehensive ability of the model.}
As we can see, Qwen2-7B improves by 3.23 points (4.7\%), Qwen1.5-14B by 2.87 points (4.1\%), Meta-Llama-3-8B by 1.23 points (2.9\%), and Mistral-Nemo-Base-2407 by 0.51 points (1.0\%), collectively demonstrating that ExMe achieves the highest overall performance across all four models. Additionally, ExMe secures the best scores in several individual metrics. For example, on the HumanEval benchmark, ExMe reaches the optimal value across all four models. Compared to the SFT models, it maintains parity with the optimal value on Qwen2-7B and improves the scores on the other three models by 0.61 points (1.0\%), 5.49 points (13.2\%), and 1.83 points (3.3\%), respectively.
\paragraph{ExMe enhances the overall performance of the model, rather than individual capabilities.}
As we can see from Table~\ref{tab:ComprehensiveComparison1} and Table~\ref{tab:ComprehensiveComparison2}, ExMe improves the overall performance of the model, but it does not achieve consistent improvements across individual capabilities, especially in areas that decline after SFT. For instance, Qwen2-7B declines by 2.33 points, and Qwen1.5-14B declines by 0.5 points in CMMLU separately. Meta-Llama-3-8B declines by 0.4 points in MMLU, and Mistral-Nemo-Base-2407 declines by 0.2 points in MATH. ExMe successfully reverses the decline in Mistral-Nemo-Base-2407 (with an increase of 0.32 points) but causes additional declines in the other three models by 0.87 points, 0.57 points, and 0.60 points, respectively. We attribute this issue to the instructional style of the SFT datasets, as the alpaca-gpt4-data-zh/en dataset is more aligned with dialogue-based question-answering, while these benchmarks mostly involve multiple-choice or multi-select questions. The lack of diversity in SFT data diminishes the model's ability to handle multiple-choice questions, and our method does not fundamentally address this issue.

\subsection{Extrapolation Stage Analysis}

\begin{figure}[!bt]
    \centering
    \includegraphics[width=1\linewidth]{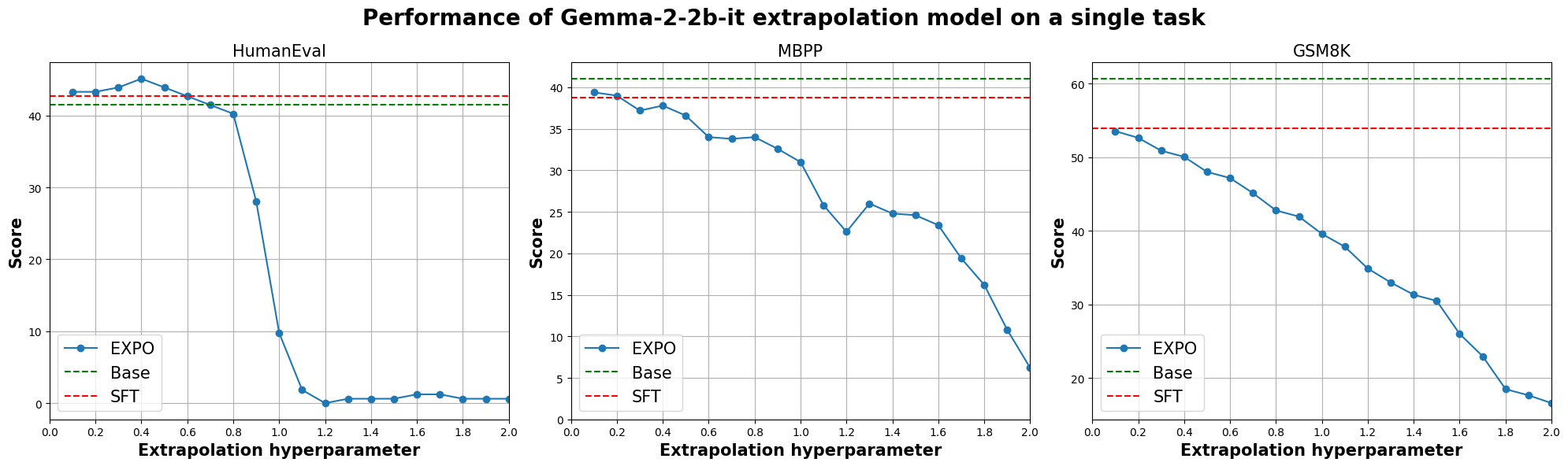}
    \caption{The extrapolation model of the gemma-2-2b-it model on HumanEval, MBPP, and GSM8K.}
    \label{fig:gemma}
\end{figure}


\begin{figure}[!bt]
    \centering
    \includegraphics[width=1\linewidth]{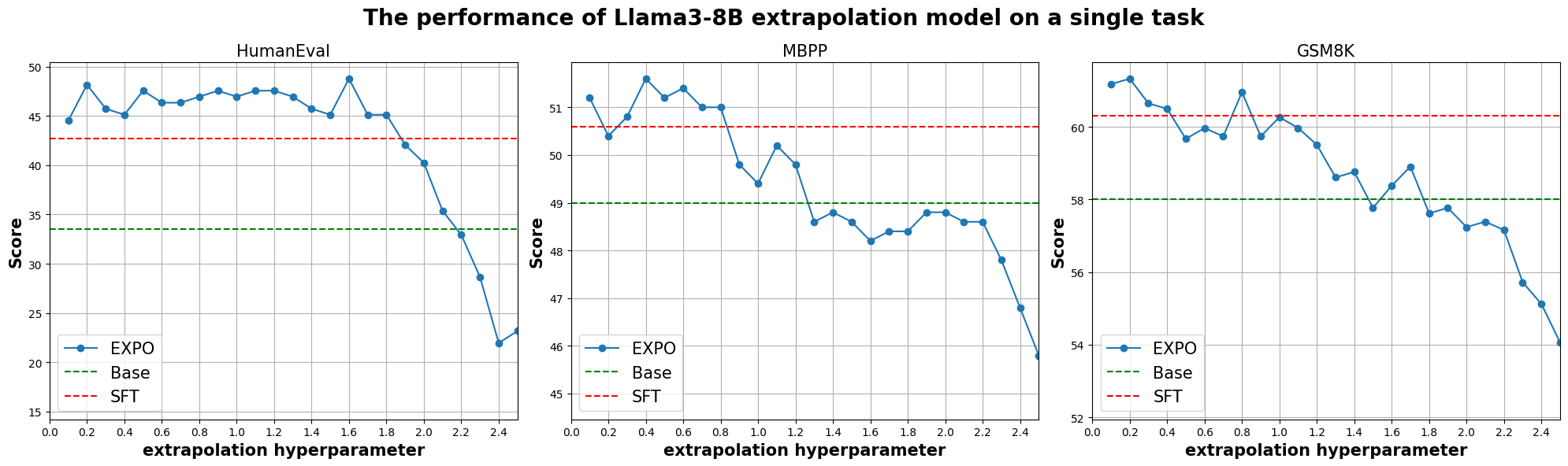}
    \caption{The extrapolation model of the Llama3-8B model on HumanEval, MBPP, and GSM8K.}
    \label{fig:llama}
\end{figure}

We validate the usefulness of the model extrapolation method in the SFT phase from the perspectives of both individual performance and overall performance. We first verify the usefulness of the model extrapolation method in the fine-tuning stage. During the fine-tuning stage, we select the checkpoints of the fine-tuning model as strong models and the base model as weak models for model extrapolation.
\paragraph{Model extrapolation is affected by parameter changes during the fine-tuning phase and has performance limits.} As shown in Figure~\ref{fig:gemma} and Figure~\ref{fig:llama}, when the performance of the SFT model is better than that of the base model, the performance of the extrapolation model shows a trend of first increasing and then decreasing with the expansion of the extrapolation coefficient. Compared with the SFT model, the extrapolation model of gemma-2-2b-it has a maximum improvement of 2.44 points on HumanEval, and the extrapolation model of Llama3-8B has a maximum improvement of 5.49 points on HumanEval, 1.00 points on MBPP, and 1.0 point on GSM8K. This indicates that the extrapolation method itself has limitations; When the performance of the SFT model is weaker than that of the base model, as shown in Figure~\ref{fig:gemma}, the performance of the extrapolation model of gemma-2-2b-it on MBPP and GSM8K always shows a downward trend with the expansion of the extrapolation coefficient, and is inferior to the two models involved in extrapolation. This indicates that extrapolation methods can to some extent expand the impact of fine-tuning on model performance, resulting in greater performance improvements or performance degradation.

\begin{figure*}[!htbp]
    \centering
    \includegraphics[width=0.9\textwidth]{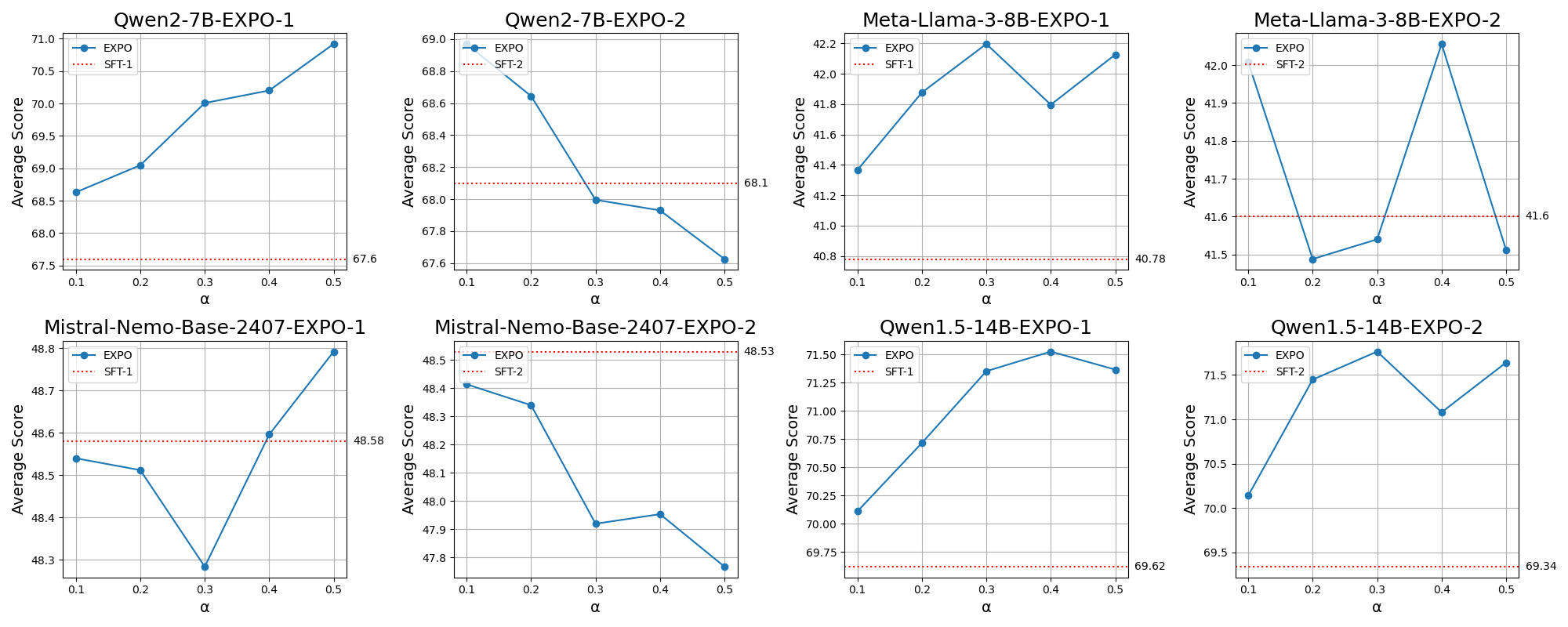}
    \caption{This figure shows the impact of the extrapolation hyperparameter $\alpha$ on the performance of the extrapolated models across four different models, where each of the eight selected SFT models is extrapolated from the base model. The blue line represents the performance of the extrapolated models, while the red dashed line indicates the performance of the SFT models involved in the extrapolation.}
    \label{fig:EXPO}
\end{figure*}

\paragraph{When considering overall performance, model extrapolation methods are also effective.}
As shown in Figure~\ref{fig:EXPO}. In the extrapolation results of the seven SFT models, we identify suitable extrapolation hyperparameters that enhance the overall performance of the extrapolated models beyond that of the original SFT models. Remarkably, for four of the SFT models, the overall performance of all extrapolated models is better than that of the original SFT model after extrapolation. The extrapolation fails only in one case of the base model Mistral-Nemo-Base-2407.

\subsection{Merging Stage Analysis}
\begin{figure}[!bt]
    \includegraphics[width=\linewidth]{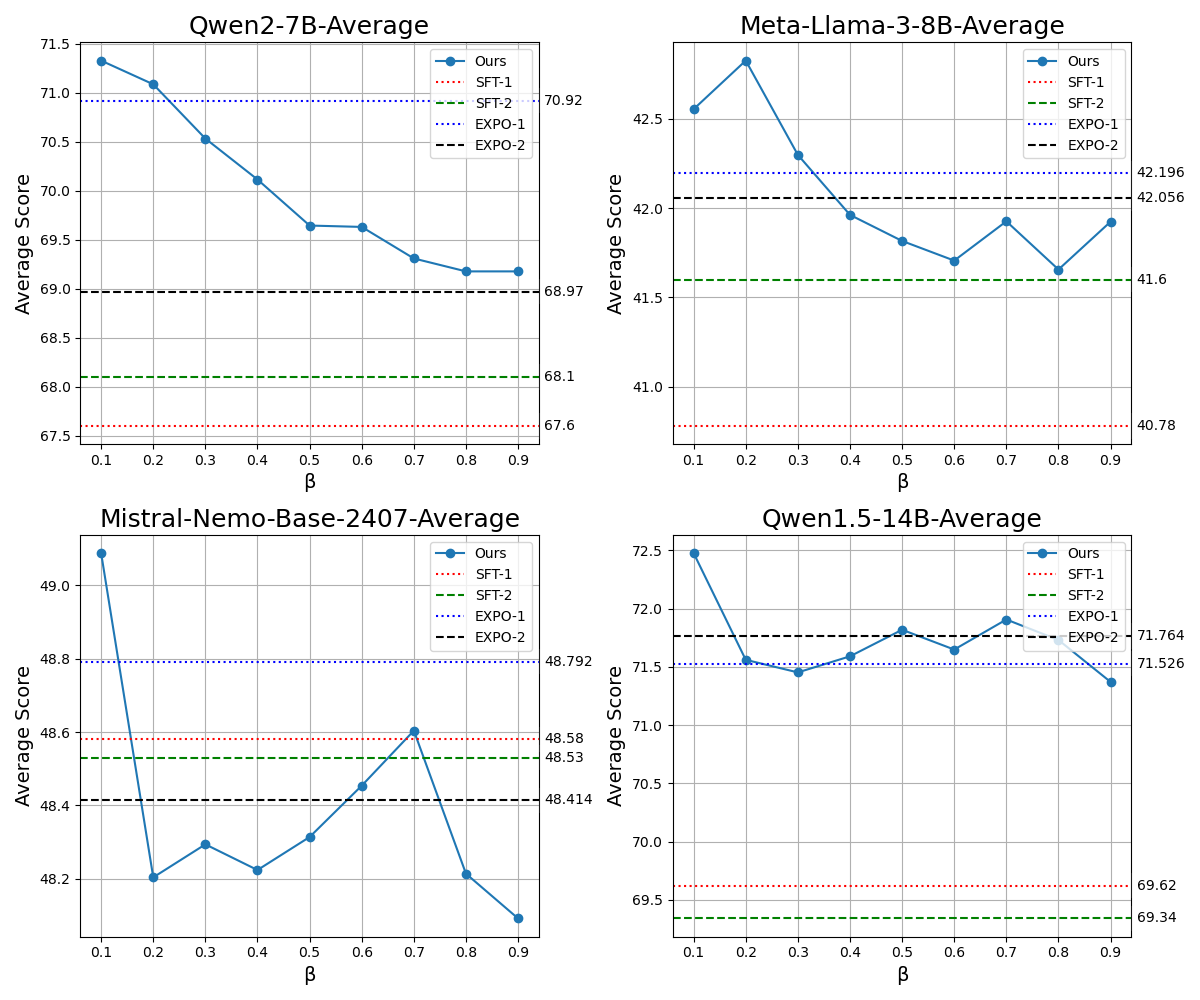}
    \caption{This figure illustrates the impact of different merging hyperparameters $\beta$ on the overall performance of candidate models across four different models, where a smaller \(\beta\) indicates a greater weight assigned to the better-performing extrapolated model during merging. The four dashed lines in the figure represent the selected models: SFT-1, SFT-2, EXPO-1, and EXPO-2.}
    \label{fig:Final}
\end{figure}
In this section, we illustrate the effectiveness of ExMe by analyzing the experimental results from the merging stage. Additionally, we examine the impact of the merging hyperparameter $\beta$ on the overall performance of all candidate models during the merging stage of ExMe, as illustrated in Figure~\ref{fig:Final}.

\paragraph{Most candidate models outperform the SFT models involved in the merging stage.} 
Except for the results of Mistral-Nemo-Base-2407 where only two candidate models ($\beta$ = 0.1 and $\beta$ = 0.7) outperform the SFT model, all other 27 candidate models in the remaining three models outperform the SFT models.
Among them, 17 candidate models exhibit an overall performance improvement exceeding 2\%, and 7 candidate models show an improvement exceeding 3\%. This indicates that ExMe offers a relatively consistent improvement in model performance.

\paragraph{Models merged with the smallest \(\beta\) value usually perform the best overall  performance.}
The optimal $\beta$ for Meta-Llama-3-8B is 0.2, while the optimal $\beta$ for the other three models is 0.1. This suggests that assigning a greater value to the merging hyperparameter for the extrapolated model with better overall performance is more likely to result in a superior model. We believe this is due to the weaker extrapolated model adjusting the parameters of the stronger extrapolated model to achieve better performance.

\paragraph{As $\beta$ increases, the model's performance shows a fluctuating downward trend.}
The performance of all four models exhibits a generally downward trend with fluctuations as $\beta$ increases. This phenomenon is particularly evident for Qwen2-7B, where the trend is clearly monotonic. We believe this is due to the obvious performance gap between the two extrapolated models being merged, with an average difference of 1.95 points, compared to differences of 0.14, 0.378, and 0.238 points for the other three models. When there is a obvious performance gap between the two models being merged, a monotonic trend is more likely to be observed. This observation is consistent with the findings in Bayesian search~\cite{DBLP:journals/corr/abs-2403-19390}.

\subsection{Comparison of ExMe with Other Methods}
We compare ExMe with other model merging methods from multiple perspectives, and our experimental results show that ExMe not only improves model performance but also offers advantages in other areas.

\paragraph{ExMe demonstrates greater robustness.} 
As shown in Table~\ref{tab:ComprehensiveComparison1} and Table~\ref{tab:ComprehensiveComparison2}, model extrapolation leads to a decrease of 3.66 points in HumanEval for Qwen1.5-14B, 1.29 points in GaoKaoBench for Meta-Llama-3-8B, 0.76 points in MATH for Mistral-Nemo-Base-2407, and 0.69 points in GaoKaoBench for the same model. In contrast, ExMe not only mitigates this downward trend but also achieves the best scores in individual tasks. Compared to the SFT models involved in the merging, it shows an improvement of 0.61 points, 0.57 points, 0.32 points, and 0.12 points, respectively.
This demonstrates that ExMe exhibits greater robustness, balancing performance improvement and the stability of the approach itself.
\paragraph{ExMe balances the strengths and weaknesses of two extrapolated models.}
From Table~\ref{tab:ComprehensiveComparison2}, we observe that for Meta-Llama-3-8B, EXPO-1 performs the best performance on MATH, while EXPO-2 excels on GSM8K. Both models demonstrate equivalent capabilities on MMLU. ExMe combines the strengths of both models, achieving balanced performance across GSM8K, MATH, and MMLU (at least stronger than the weaker of the two extrapolated models). Notably, for HumanEval and GaoKaoBench, ExMe surpasses the limits of individual model extrapolation, improving the scores by 1.22 and 1.7 points, respectively. Similarly, this pattern manifests in other models.

\section{Conclusion}
In this work, we apply model extrapolation to the SFT stage, we find that model extrapolation can leverages base models and SFT models to create better-performing models. In addition, we explore how different extrapolation hyperparameters affect the performance of the extrapolated models. 

Furthermore, building on the effectiveness of model extrapolation, we propose a simple and effective model merging strategy called ExMe. ExMe provides a clear direction for model merging based on extrapolation in SFT stage and can reliably enhance model performance after SFT without requiring any additional training.

Our experiments validate the effectiveness of ExMe, demonstrating that it provides greater improvements in overall model performance and robustness compared to other merging method.

\section{Limitation}
Our work primarily relies on model merging and model extrapolation methods, both of which lack rigorous mathematical proofs. While we have experimentally validated the effectiveness of these methods, mathematical proofs have not been provided. In our approach, both the model extrapolation and merging phases involve all parameters in the process, without identifying or addressing the impact of redundant parameters. This represents a potential direction for future optimization.

\bibliography{custom}

\appendix

\label{sec:appendix}

\end{document}